# Large Language Models Versus Physicians in Traditional Chinese Medicine: A Real-World Clinical Case Evaluation

Jiacheng Xie[1,2†], Xiaoting Tang[3†], Yang Yu[1,2], Jinpu Li[2], Shouli Li[4], Congcong Jing[5], Yantao Yang[6], Zhiyong Zhao[7], Ziyang Zhang[8], Qilin Song[9], Guanghui An[1,2,10]*, Dong Xu[1,2]*

[1] Department of Electrical Engineering and Computer Science, University of Missouri, Columbia, MO, USA;
[2] Christopher S. Bond Life Sciences Center, University of Missouri, Columbia, MO, USA;
[3] Community Health Service Center Shanghai Pudong New Area, Shanghai, China;
[4] Jinsha County Chinese Medicine Hospital, Bijie, Guizhou Province, China;
[5] Shanghai Seventh People's Hospital, Shanghai, China;
[6] The First People's Hospital of Lanzhou City, Lanzhou, Gansu Province, China;
[7] The First Affiliated Hospital of Yunnan University of Chinese Medicine, Kunming, Yunnan Province, China;
[8] Department of Computer Science, Northwestern University, Chicago, IL, USA;
[9] Linyi Traditional Chinese Medicine Hospital, Linyi, Shandong Province, China;
[10] School of Acupuncture-Moxibustion and Tuina, Shanghai University of Traditional Chinese Medicine, Shanghai, China.

[†] These authors contributed equally: Jiacheng Xie, Xiaoting Tang.

* Corresponding authors:
Guanghui An, E-mail: agh@shutcm.edu.cn
Dong Xu, E-mail: dxu1@usf.edu

**Abstract**

Large language models (LLMs) are increasingly being explored for clinical applications, yet their assessment for real-world traditional Chinese medicine (TCM) practice remains limited. We constructed a clinical case library comprising 349 de-identified outpatient cases from 62 hospitals and evaluated 16 LLMs and a comparator cohort of 60 practicing TCM physicians using 60 representative cases selected from this library. Model outputs and physician reports were anonymized and scored by five senior TCM experts across nine diagnostic and therapeutic dimensions. Cutting-edge general-purpose LLMs achieved higher expert scores than the physician comparators, particularly for medical advice, treatment principles and selected diagnostic tasks. However, prescription-level analyses revealed discrepancies in herb selection, dosage, and treatment strategy, and qualitative safety review identified hallucinations and undesirable template-driven outputs. These findings highlight the potential of LLMs for TCM decision support while underscoring the need for physician oversight, safety constraints and prospective clinical evaluation.

**Introduction**

Patients with complex or unresolved clinical presentations remain difficult to manage in routine care. Many undergo prolonged diagnostic pathways, repeated investigations and periods of uncertainty before an appropriate diagnosis or treatment plan is reached[1,2]. These cases place a substantial burden on clinicians, who must integrate incomplete histories, diverse clinical findings and a growing medical literature, often within limited consultation time[3,4]. Large language models (LLMs) have attracted interest in this context because they can process free-text clinical information, summarize dispersed knowledge and generate case-specific responses[3–5]. Studies have shown that LLMs can perform well on medical examinations and clinical knowledge benchmarks, including the United States Medical Licensing Examination and MultiMedQA[6,7]. Further work has reported improved performance in long-form medical question answering and health-system-scale prediction tasks, suggesting possible roles for language models in selected clinical workflows[8,9].

Whether these systems remain clinically useful and safe in real cases is less clear. A recent systematic review identified 4609 peer-reviewed studies of LLMs in clinical medicine between January 2022 and September 2025, but only 1048 used real-world patient data and only 19 were prospective randomized trials[10]. Many evaluations still rely on examination questions, simulated vignettes or expert-written prompts rather than tasks arising from routine care[10,11]. This distinction is important. In real clinical settings, information is often incomplete, decisions are made step by step, and an apparently plausible response may still be unsafe or clinically inappropriate[11,12]. Recent studies have begun to evaluate LLMs in patient communication, diagnostic reasoning, emergency decision support and clinical documentation workflows[13–17]. These studies suggest potential clinical value but also highlight the need for careful evaluation. LLMs may generate unsupported statements, omit relevant findings, fail to follow instructions or provide recommendations that conflict with accepted standards of care[12,18,19]. Current frameworks for evaluating clinical AI therefore emphasize transparent study design, clinically appropriate comparators, expert review and explicit assessment of risk[20–22].

Most medical LLM evaluations have been conducted in Western medical settings and English-language datasets[3,7,10]. Other medical systems have received less attention. Traditional medicine remains part of health care in many regions, and international health policy has called for evidence-based evaluation and safe integration of traditional, complementary and integrative medicine[23,24]. Traditional Chinese Medicine (TCM), practiced across China and in many other regions, provides a clinically relevant context for evaluating medical reasoning. TCM diagnosis is based on *bianzheng*[25], a framework of pattern differentiation followed by personalized treatment determination. Physicians synthesize information from inspection, auscultation and olfaction, inquiry and palpation to identify disease patterns and guide individualized treatment[26–28]. In this setting, diagnosis and treatment are closely linked. Syndrome differentiation informs prescription composition, dosage allocation and therapeutic priorities. Clinical reasoning therefore requires not only a diagnostic label, but also a treatment plan that is coherent with the symptom narrative, tongue and pulse findings, disease course, treatment history, herbal compatibility and dosage[26–29].

Although interest in LLMs for TCM is increasing, evidence from real clinical practice remains limited. Recent studies and benchmarks have evaluated LLMs in TCM diagnosis, treatment recommendation, syndrome differentiation and prescription review[30,31]. However, most of these evaluations use examination-style questions, constructed cases, isolated examples or narrowly defined benchmark tasks rather than real outpatient encounters with practicing physician comparators[30–32]. Existing TCM benchmarks have supported structured question answering, syndrome reasoning and prescription audit, but they do not fully address how LLM-generated clinical reports compare with physician-generated reports in real-world outpatient care[31,32]. This gap is clinically important because errors in TCM are not limited to misclassification. Incorrect syndrome differentiation, unsupported herb suggestions, unsafe herb combinations or inappropriate dosage recommendations may affect prescription safety and treatment suitability[31,32].

In this study, we evaluated LLMs in real-world TCM diagnosis and prescription generation. We curated 349 authentic outpatient cases and selected 60 representative cases for structured head-to-head evaluation. We compared 16 contemporary LLMs with 60 practicing TCM physicians. All model- and physician-generated reports were anonymized and independently reviewed by senior TCM experts across diagnostic and therapeutic dimensions. We also examined prescription patterns, hallucinations, efficiency and safety-related risks. By combining real outpatient data, physician comparators, blinded expert review and prescription-level analysis, this study assesses LLM performance across the practical sequence of TCM care, from case interpretation to treatment planning. The findings provide evidence for the use and limitations of LLMs in TCM, and inform how clinical AI systems should be evaluated before they are used as decision-support tools in knowledge-intensive medical settings.

## Results

### Clinical patient dataset and evaluation framework

We constructed a real-world clinical outpatient dataset spanning 2018 to 2023, comprising 349 clinical cases collected from 62 hospitals (**Table 1**). All cases were obtained with informed consent from both patients and physicians and are released to the research community for the first time to support this evaluation. The dataset contains comprehensive patient-level information, including gender, age, occupation, date of first visit, chief complaint, history of present illness, and current condition covering the four TCM diagnostic methods (inspection, auscultation/olfaction, inquiry, and palpation), as well as tongue and pulse assessments. It also includes specialized examinations such as routine laboratory and imaging tests (e.g., complete blood counts, urinalysis, ultrasound, and electrocardiography). We further selected 60 clinically diverse and representative cases from this dataset to support the evaluation conducted in this study. All personal identifiers of both patients and physicians were removed through standardized de-identification procedures. An overview of the dataset and representative case samples is presented in **Fig. 1A**.

As shown in **Fig. 1B**, our evaluation framework comprises three components: a cohort of human clinicians, a panel of LLMs, and an independent expert scoring committee. The clinician cohort consists of 60 licensed TCM physicians recruited from 47 hospitals across China, spanning 21 clinical departments. These physicians have an average of 13.5 years of practice, and 57 hold junior or higher professional titles, ensuring a representative and experienced clinical baseline. The LLM panel includes 16 state-of-the-art models available as of 1 September 2025, comprising 12 general-purpose LLMs and 4 TCM-specialized models. The general-purpose models include GPT-5[33], GPT-4o[34], and GPT-o3[35]; Gemini 2.5 Pro[36]; Grok 3[37] and Grok 4[38];

Claude Opus 4[39]; Llama-4-Maverick[40]; Doubao 1.6 and Doubao 1.6 Thinking[41]; as well as Qwen 3-256B[42] and DeepSeek-R1[43]. For domain-specific modeling, we reviewed 11 published TCM-domain LLMs and selected four openly available models for systematic evaluation: BianCang[44], HuaTuoGPT2[45], BenCao[46], and Huatuo[47]. Among these, BenCao is a TCM-specialized model developed by our team through instruction tuning on ChatGPT to better accommodate clinically oriented TCM reasoning tasks. The expert scoring committee consists of five senior TCM specialists who were blinded to all model and clinician identities and did not participate in case diagnosis. They provided an independent blinded assessment of diagnostic quality.

To approximate real clinical conditions, participating clinicians were allowed to select cases aligned with their subspecialty expertise. Based on these preferences, we assigned cases such that each clinical case was evaluated by six physicians and each physician diagnosed six cases as shown in **Supplementary Table 1**. Before the evaluation, both the LLMs and clinicians were given an identical example of the required diagnostic format and instructed to strictly adhere to this structure when generating their reports. All LLMs evaluated the complete benchmark cohort of 60 cases, whereas each physician assessed only the cases assigned to them. For each case, diagnostic reports generated by clinicians and LLMs were pooled and randomly mixed. The expert committee scored these reports in a single-blind manner using a five-point Likert scale according to a unified rubric, as described in **Supplementary Table 2**. To facilitate efficient and standardized scoring, we developed a dedicated online assessment platform that supported blinded review and automatically recorded all evaluation data. Further details regarding model selection and the full scoring procedure are provided in the **Methods section**.

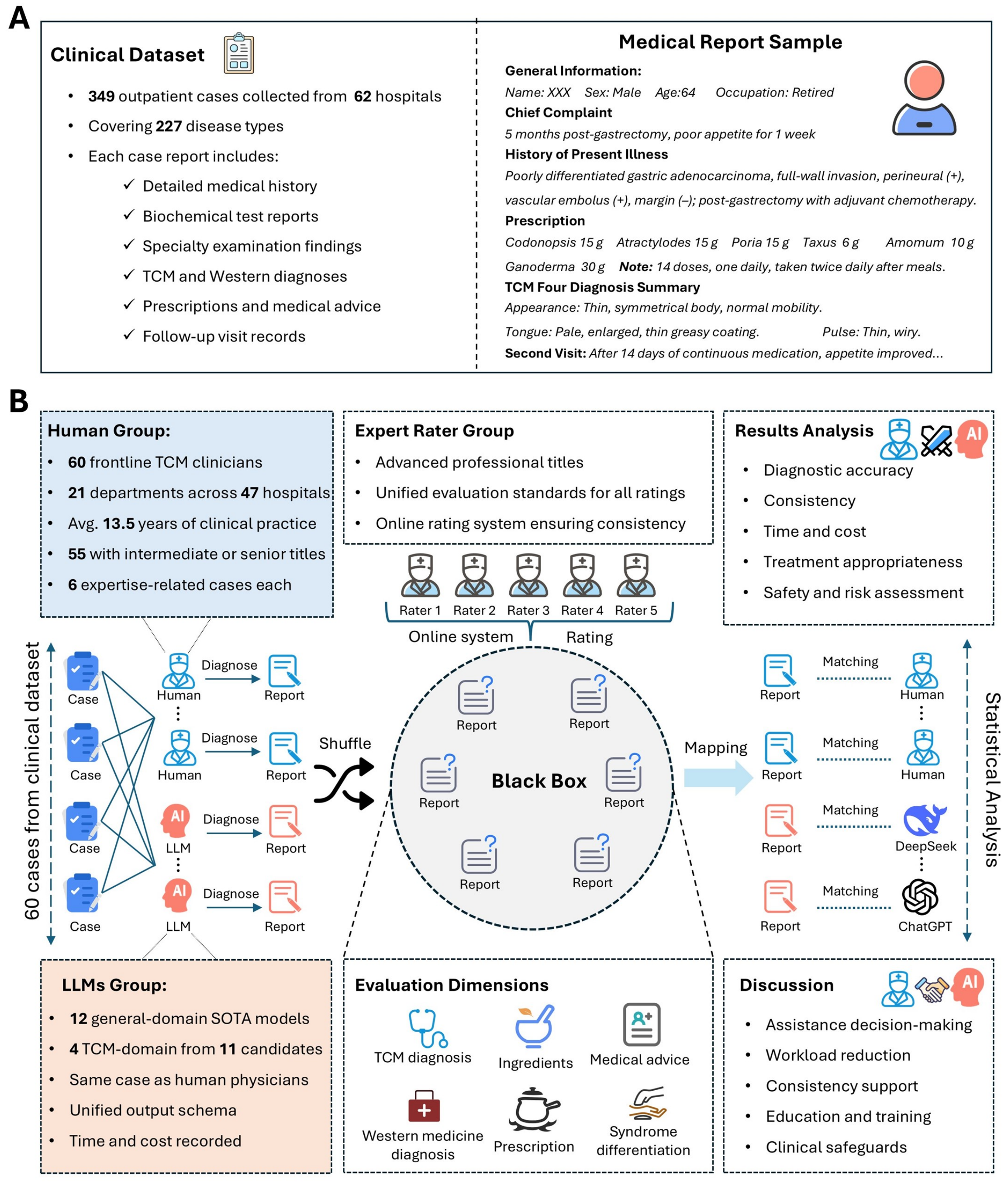

**Fig. 1 | Framework for clinical dataset construction and comparative evaluation of LLMs and physicians. (A)** A clinical dataset comprising 349 outpatient cases from 62 hospitals, covering 227 disease types, was curated. Each case included detailed medical history, biochemical tests, specialty examinations, TCM and Western diagnoses, prescriptions, medical advice, and follow-up visit records. A structured medical report sample illustrates the standardized format. **(B)** 60 frontline TCM physicians from 21 departments across 47 hospitals and 16 LLMs (12 general-domain and 4 TCM-domain SOTA models) diagnosed the same cases

under identical conditions. Each physician selected six expertise-related cases. Diagnostic reports were independently evaluated by an expert rater group using a unified online scoring system. Results were analyzed across multiple evaluation dimensions, including diagnostic accuracy, consistency, efficiency, treatment appropriateness, and safety. Discussion highlights both the potential of LLMs to complement human expertise in clinical decision-making and the importance of safeguards for reliable clinical deployment.

**Table 1. Data reported as numbers of patients, with percentages in parentheses**

| ***Variable*** | *Source cohort* | *Benchmark dataset* |
| --- | --- | --- |
| *Total number of cases* | 349 | 60 |
| ***Sex*** | | |
| *Female* | 225 (64.5%) | 41 (68.3%) |
| *Male* | 120 (34.4%) | 19 (31.7%) |
| ***Age*** | | |
| *Mean age* | 49.2 years | 55.5 years |
| *Median age* | 52.0 years | 56 years |
| *Standard deviation* | 19.1 years | 16.2 years |
| *Minimum age* | 3 years | 26 years |
| *Maximum age* | 92 years | 83 years |
| *<40 years* | 94 (27.6%) | 12 (20.0%) |
| *40-49 years* | 65 (19.1%) | 12 (20.0%) |
| *50-59 years* | 63 (18.5%) | 9 (15.0%) |
| *60-69 years* | 70 (20.5%) | 13 (21.7%) |
| *≥70 years* | 49 (14.4%) | 14 (23.3%) |
| ***Case provenance*** | | |
| *Unique physicians* | 75 | 42 |
| *Initial date span* | 2018-2023 | 2021-2023 |
| ***Disease category*** | | |
| *Renal / urinary* | 109 (31.2%) | 10 (16.7%) |
| *Thyroid-related* | 10 (2.9%) | 9 (15.0%) |
| *Gynecological / reproductive* | 16 (4.6%) | 6 (10.0%) |
| *Diabetes / metabolic* | 4 (1.1%) | 6 (10.0%) |
| *Cardiovascular* | 17 (4.9%) | 6 (10.0%) |
| *Respiratory* | 19 (5.4%) | 5 (8.3%) |
| *Liver / cirrhosis-related* | 2 (0.6%) | 5 (8.3%) |
| *Digestive* | 28 (8.0%) | 4 (6.7%) |
| *Dizziness / vertigo-related* | 5 (1.4%) | 3 (5.0%) |
| *Tumor-related conditions* | 35 (10.0%) | 3 (5.0%) |
| *Benign breast* | 5 (1.4%) | 2 (3.3%) |
| *Dermatological* | 22 (6.3%) | 1 (1.7%) |
| *Anorectal* | 3 (0.9%) | 0 (0.0%) |
| *ENT / oral* | 9 (2.6%) | 0 (0.0%) |
| *Hematologic / immune / rheumatologic* | 3 (0.9%) | 0 (0.0%) |
| *Musculoskeletal / pain* | 29 (8.3%) | 0 (0.0%) |
| *Neurologic / psychiatric / sleep* | 27 (7.7%) | 0 (0.0%) |
| *Other / mixed* | 6 (1.7%) | 0 (0.0%) |

**Cutting-edge general-domain LLMs achieved higher expert scores than the physician comparator across multiple clinical evaluation dimensions**

As shown in **Fig. 2A**, several cutting-edge general-purpose LLMs achieved higher blinded expert scores than the physician comparator across multiple dimensions in this structured case evaluation. Among the evaluated models, DeepSeek-R1 achieved the highest overall performance, exceeding the physician baseline by 0.667 points on average, followed by Claude Opus 4 and GPT-5, with average differences of 0.629 and 0.627, respectively. Notably, several cutting-edge general-purpose models exhibit positive deviations across nearly all evaluation dimensions, indicating robust and generalizable diagnostic capabilities. Performance gains were most pronounced in Medical Advice and Usage, where several models exceeded the physician baseline by more than 1 point. Gains in Treatment Principle were also positive for many cutting-edge models, although smaller in magnitude. These findings suggest that LLMs are particularly effective in synthesizing clinical information into actionable recommendations. In addition, strong performance is observed in core TCM-related dimensions such as *Syndrome differentiation*, *Prescription*, and *Ingredients*, where leading models achieve parity with or surpass clinicians. However, performance was highly heterogeneous across models and evaluation dimensions. While most cutting-edge general-purpose LLMs outperformed the human baseline across multiple categories, several lower-performing models, particularly HuatuoGPT-2, BianCang, and Huatuo, remained substantially below physicians overall. Their deficits were most evident in Syndrome, Prescription, Western Diagnosis, Treatment Principle, and TCM Diagnosis, although some isolated strengths were observed in Usage. Notably, BenCao, a TCM-specialized model developed through instruction tuning of GPT-4o, consistently outperformed both GPT-4o and the human baseline across multiple dimensions, indicating that targeted adaptation of a strong general-purpose foundation model can yield measurable gains in TCM-oriented clinical reasoning. However, its overall performance remained below that of the strongest cutting-edge models, particularly GPT-5, DeepSeek-R1, and Claude Opus 4.

**Fig. 2B** further supports these findings at the distributional level. Across multiple evaluation dimensions, cutting-edge general-purpose LLMs generally showed right-shifted score distributions and higher median scores than the physician baseline. This pattern was especially apparent in treatment-oriented dimensions, where the score distributions of leading models were concentrated toward higher values. By contrast, lower-performing models showed lower overall score distributions and, in several dimensions, greater dispersion across cases, indicating less stable performance across the clinical case set.

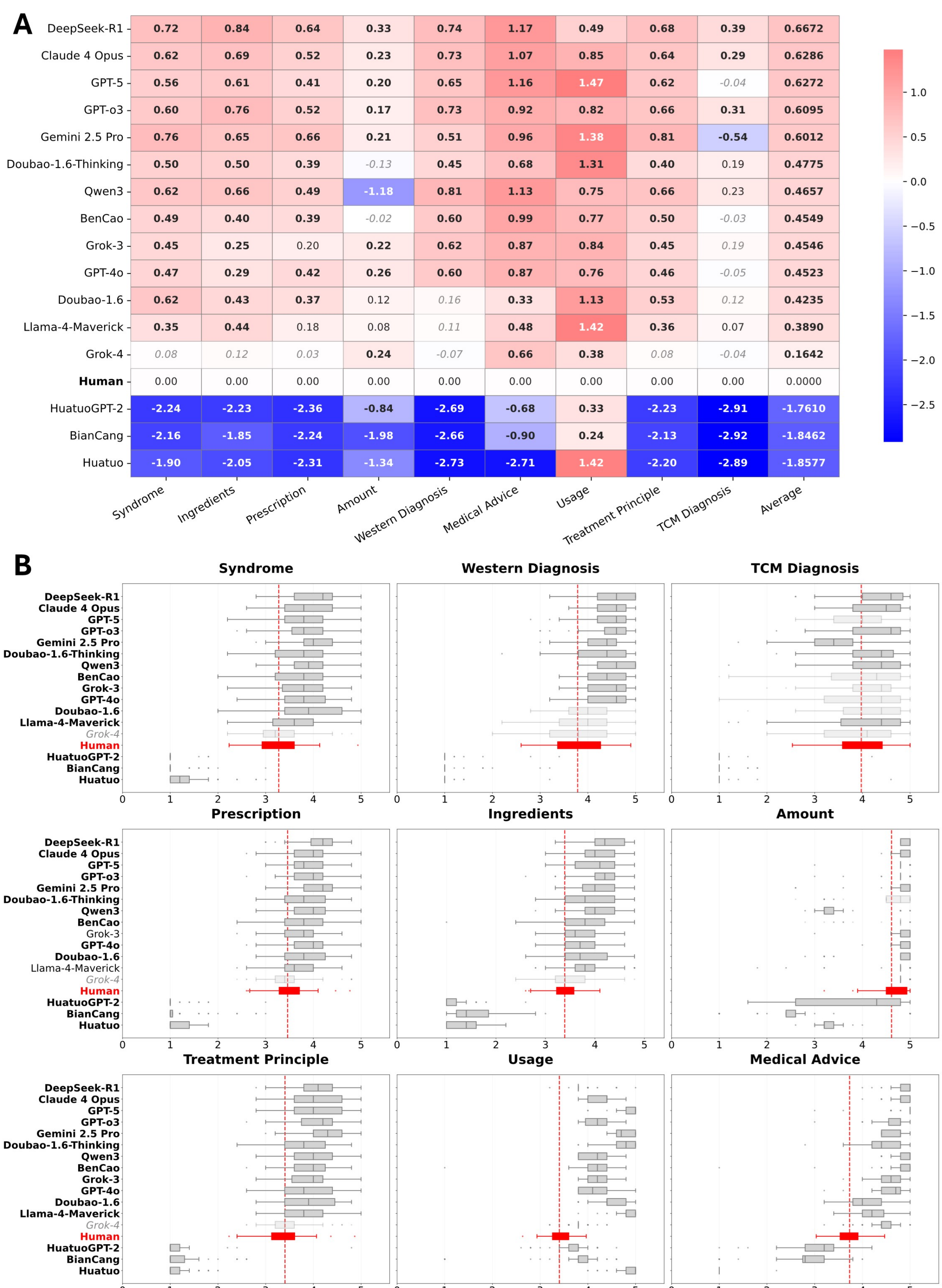
A
DeepSeek-R1 0.72 0.84 0.64 0.33 0.74 1.17 0.49 0.68 0.39 0.6672
Claude 4 Opus 0.62 0.69 0.52 0.23 0.73 1.07 0.85 0.64 0.29 0.6286
GPT-5 0.56 0.61 0.41 0.20 0.65 1.16 1.47 0.62 -0.04 0.6272
GPT-o3 0.60 0.76 0.52 0.17 0.73 0.92 0.82 0.66 0.31 0.6095
Gemini 2.5 Pro 0.76 0.65 0.66 0.21 0.51 0.96 1.38 0.81 -0.54 0.6012
Doubao-1.6-Thinking 0.50 0.50 0.39 -0.13 0.45 0.68 1.31 0.40 0.19 0.4775
Qwen3 0.62 0.66 0.49 -1.18 0.81 1.13 0.75 0.66 0.23 0.4657
BenCao 0.49 0.40 0.39 -0.02 0.60 0.99 0.77 0.50 -0.03 0.4549
Grok-3 0.45 0.25 0.20 0.22 0.62 0.87 0.84 0.45 0.19 0.4546
GPT-4o 0.47 0.29 0.42 0.26 0.60 0.87 0.76 0.46 -0.05 0.4523
Doubao-1.6 0.62 0.43 0.37 0.12 0.16 0.33 1.13 0.53 0.12 0.4235
Llama-4-Maverick 0.35 0.44 0.18 0.08 0.11 0.48 1.42 0.36 0.07 0.3890
Grok-4 0.08 0.12 0.03 0.24 -0.07 0.66 0.38 0.08 -0.04 0.1642
Human 0.00 0.00 0.00 0.00 0.00 0.00 0.00 0.00 0.00 0.0000
HuatuoGPT-2 -2.24 -2.23 -2.36 -0.84 -2.69 -0.68 0.33 -2.23 -2.91 -1.7610
BianCang -2.16 -1.85 -2.24 -1.98 -2.66 -0.90 0.24 -2.13 -2.92 -1.8462
Huatuo -1.90 -2.05 -2.31 -1.34 -2.73 -2.71 1.42 -2.20 -2.89 -1.8577
Syndrome
Ingredients
Prescription
Amount
Western Diagnosis
Medical Advice
Usage
Treatment Principle
TCM Diagnosis
Average
1.0
0.5
0.0
−0.5
−1.0
−1.5
−2.0
−2.5
B
Syndrome
Western Diagnosis
TCM Diagnosis
Prescription
Ingredients
Amount
Treatment Principle
Usage
Medical Advice
DeepSeek-R1
Claude 4 Opus
GPT-5
GPT-o3
Gemini 2.5 Pro
Doubao-1.6-Thinking
Qwen3
BenCao
Grok-3
GPT-4o
Doubao-1.6
Llama-4-Maverick
Grok-4
Human
HuatuoGPT-2
BianCang
Huatuo
0 1 2 3 4 5
Score

**Fig. 2 | Diagnostic performance of LLMs relative to physicians across nine clinical reasoning dimensions. (A)** Heatmap of the mean score difference (Δ) between each model and the physician baseline, with the physician baseline set to 0. Scores were aggregated across $n = 60$ clinical cases and five expert raters. Nine evaluation dimensions were included: Syndrome, Ingredients, Prescription, Amount, Western Diagnosis, Medical Advice, Usage, Treatment Principle, and TCM Diagnosis. Red indicates higher scores than the physician baseline, whereas blue indicates lower scores. Statistical significance was assessed using two-sided paired Wilcoxon signed-rank tests, followed by Benjamini–Hochberg false discovery rate correction. Bold values indicate BH-adjusted $P < 0.001$, and faded italic values indicate non-significant comparisons with BH-adjusted $P \geq 0.05$. The Average column reports the mean difference across all nine dimensions. **(B)** Box plots showing per-case score distributions for each model across the same nine dimensions. Each distribution contains $n = 60$ cases. Boxes represent the interquartile range, center lines indicate medians, whiskers extend to 1.5 times the interquartile range, and points denote outliers. The red box denotes the physician baseline distribution, calculated from physician scores aggregated at the case level. The red dashed vertical line indicates the mean physician score for each dimension.

**Performance of TCM-Specialized LLMs falls short of general-domain LLMs in case-level ranking and win–tie–loss analyses**

**Fig. 3** provides complementary evidence for the performance advantage of cutting-edge LLMs over physicians at both the ranking and case levels. In the average-rank comparison as shown in **Fig. 3A**, cutting-edge general-purpose models occupied the top positions, with DeepSeek-R1 achieving the best overall rank (4.34), followed by GPT-5 (4.83), Claude 4 Opus (4.92), GPT-o3 (4.96), and Gemini 2.5 Pro (5.28). By contrast, the human average ranked substantially lower (13.18), below nearly all general-purpose LLMs and above only the lowest-performing TCM-oriented models. The Friedman test indicated highly significant overall differences across models ($p = 2.27 \times 10^{-107}$), and the critical-difference diagram showed that several of the top-ranked cutting-edge models were not significantly different from one another, indicating a shared leading performance tier. In contrast, HuatuoGPT-2, HuaTuo, and BianCang occupied the lowest ranks, highlighting the limited competitiveness of these domain-specific models under the present evaluation setting. The win–tie–loss analysis against physicians further reinforced this pattern as shown in **Fig. 3B**. The strongest models outperformed physicians in the large majority of cases, with DeepSeek-R1 winning in 58/60 cases, Claude Opus 4 and Gemini 2.5 Pro in 57/60, and GPT-5 and GPT-o3 in 55/60. Several mid-tier models also showed favorable but less dominant performance, including Grok 3 (53/60 wins), Qwen3-256B (51/60), GPT-4o (50/60), and Doubao-1.6-Thinking (49/60). BenCao outperformed physicians in 46/60 cases, exceeding its base model GPT-4o in average rank but remaining below the strongest cutting-edge systems overall. By contrast, the lowest-performing TCM-oriented models registered virtually no wins against physicians, underscoring their limited practical competitiveness in this benchmark.

**Case-level analyses reveal LLM advantages across challenging clinical scenarios**

The case-difficulty analysis provided additional insight into these performance differences, as shown in **Fig. 3C**. When average physician scores were plotted against average LLM scores for each of the 60 clinical cases, the lower-left region corresponded to cases that were difficult for both groups, whereas the upper-right region corresponded to cases that were relatively easy for both. Notably, harder cases tended to fall above the diagonal, indicating that LLMs often retained a relative advantage in more challenging scenarios. By contrast, easier cases more often clustered near or below the diagonal, suggesting comparatively stronger human performance on simpler cases. These results indicate that the relative advantage of cutting-edge LLMs was not limited to straightforward cases, but was maintained, and in some instances amplified, under more difficult clinical conditions. Case-level comparisons across all 60 clinical cases showed a consistent but non-uniform pattern as shown in **Fig. 3D**. Cutting-edge models typically outperformed five to six physicians per case, with the highest average counts observed for DeepSeek-R1, Claude 4 Opus, and Gemini 2.5 Pro (all 5.7), followed by GPT-o3 (5.6) and GPT-5 (5.5). Mid-tier models such as Qwen3-256B, GPT-4o, BenCao, and Grok 3 remained competitive but exhibited greater case-to-case variation. This heterogeneity indicates that model advantages were robust overall but not uniform across clinical scenarios. In contrast, the weakest TCM-oriented models rarely outperformed individual physicians at the case level, further confirming a substantial gap between cutting-edge general-purpose LLMs and lower-performing domain-specific systems.

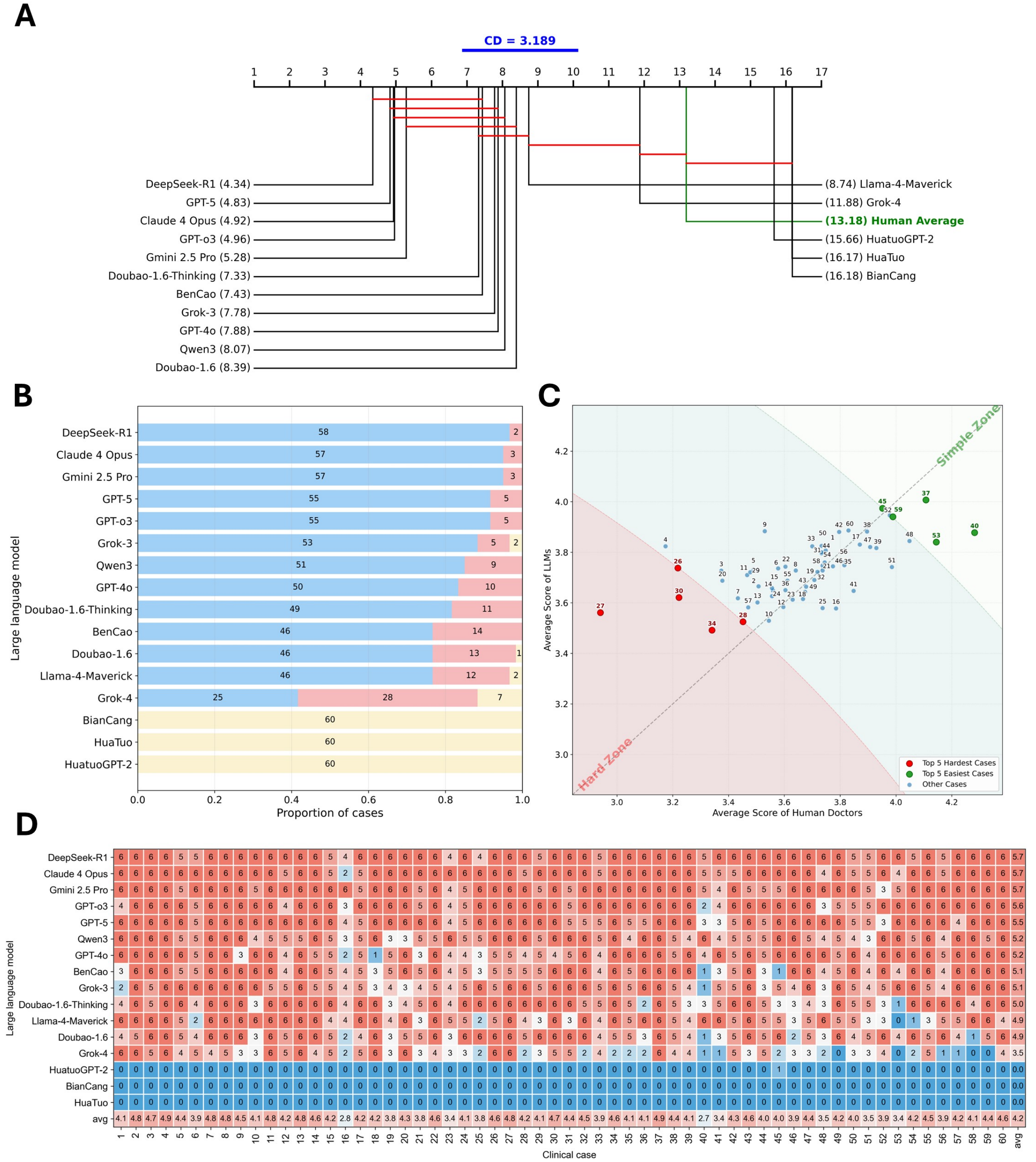

**Fig. 3 | Case-level ranking and win–tie–loss analyses comparing LLMs with physicians. (A)** Critical difference diagram showing the average ranks of all models and the human average across 60 clinical cases, with lower ranks indicating better overall performance. Statistical significance was assessed using the Friedman test ($P = 2.27 \times 10^{-107}$), and the critical difference (CD = 3.189) denotes the threshold for significant pairwise differences. Models connected by horizontal red bars are not significantly different from one another. Cutting-edge general-purpose LLMs occupy the top ranks, whereas the human average ranks lower than most general-purpose

models and above only the lowest-performing TCM-oriented models. **(B)** Case-level win–tie–loss comparison of each model against the physician reference across the same 60 cases, shown as proportions of cases. Cutting-edge LLMs achieve wins in most cases, with the top models exceeding 90% win rates, whereas lower-performing TCM-oriented models rarely or never win. **(C)** Scatter plot comparing the average score of physicians (x axis) and the average score of LLMs (y axis) for each clinical case, providing a case-level view of task difficulty. Cases in the lower-left region represent those that are difficult for both humans and LLMs, whereas cases in the upper-right region represent those that are relatively easy for both. The dashed diagonal indicates equal average performance between the two groups. Red points denote the five hardest cases, green points denote the five easiest cases, and blue points denote all remaining cases. Harder cases tend to lie above the diagonal, indicating that LLMs often retain a relative advantage on more difficult cases, whereas easier cases more often fall near or below the diagonal, suggesting stronger human performance on simpler cases. **(D)** Heat map showing, for each clinical case, the number of physicians outperformed by each model (range, 0–6). The rightmost column indicates the average number of physicians outperformed across all cases for each model, and the bottom row indicates the average across models for each case. Cutting-edge LLMs outperform physicians in most cases, although performance varies across cases, whereas lower-performing TCM-oriented models rarely exceed individual physicians.

**DeepSeek assistance increased junior physicians' diagnostic scores**

We further evaluated whether DeepSeek-generated diagnostic support improved junior physicians' diagnostic performance in an exploratory analysis of clinician–AI interactions. Seven junior physicians independently reviewed the benchmark cases before and after access to DeepSeek-generated diagnostic support, and their diagnostic reports were anonymized and incorporated into the same blinded expert-evaluation workflow used for the model and physician outputs. All reports were scored by the same panel of five senior TCM experts using the predefined multidimensional scoring rubric.

Compared with their unaided baseline performance, DeepSeek-assisted junior physicians showed a mean increase of 0.836 points in normalized overall score. Individual-level improvements were directionally consistent across physicians, with a variance of 0.048 and a standard deviation of 0.219 in score improvement. A paired comparison indicated a statistically significant improvement after DeepSeek assistance (two-sided Wilcoxon signed-rank test, $P = 0.016$).

**LLMs generate prescriptions with more ingredients and dosages than physicians**

As shown in **Fig. 4A**, substantial model-level deviations persisted in prescription structure compared with physicians. Most cutting-edge general-purpose LLMs were shifted into the upper-right quadrant, indicating a tendency to generate prescriptions with both more ingredients and higher total dosage than the physician case mean. Models such as GPT-5, GPT-o3, and Claude Opus 4 exhibited particularly large positive deviations along one or both axes, whereas a few models, including GPT-4o and Llama-4-Maverick, remained closer to the physician reference. Among TCM-oriented models, BenCao was relatively close to the origin, whereas HuatuoGPT-2 showed a pronounced deviation toward prescriptions with fewer ingredients and markedly lower total dosage, indicating substantial under-prescription relative to physicians. At the herb-composition level, **Fig. 4B** further revealed systematic differences in prescription preferences between physicians and LLMs. LLM-generated prescriptions more frequently included herbs such as *licorice root* (Δ = +0.264), *codonopsis root* (Δ = +0.216), and *poria* (Δ = +0.175), with additional positive shifts observed for *salvia root*, *ophiopogon root*, *curcuma tuber*, and *ziziphus seed*. By contrast, physicians showed slightly higher prevalence for herbs such as aconite slice, chicken gizzard lining, dried ginger, and prepared rehmannia, although these differences were smaller in magnitude. These patterns indicate that LLMs and physicians did not merely differ in total prescription size, but also in the internal composition of formulas.

Dose-level comparisons of shared herbs showed similarly structured discrepancies. In **Fig. 4C**, among herbs prescribed by both groups in at least 10 shared cases, LLMs assigned substantially higher doses to *Bupleuri Radix* (+7.56 g, $n = 37$), with smaller positive deviations for *Eucommiae Cortex* (+1.37 g, $n = 14$) and *Zingiberis Rhizoma Recens* (+0.63 g, $n = 21$). In contrast, physicians prescribed higher mean doses for several commonly shared herbs, including *Polygoni Multiflori Caulis* (−4.27 g, $n = 13$), *Dioscoreae Rhizoma* (−3.29 g, $n = 22$), *Coicis Semen* (−2.93 g, $n = 14$), *Salviae Miltiorrhizae Radix* (−2.78 g, $n = 32$), and *Rehmanniae Radix* (−2.75 g, $n = 35$). Thus, divergence between LLMs and physicians extended beyond herb selection to clinically meaningful differences in dose calibration. These prescription-level discrepancies were further reflected in agreement metrics across diagnostic and therapeutic categories. As shown in **Fig. 4D**, cross-group alignment between LLMs and physicians was generally lower than within-group consistency for at least one group across all nine evaluated dimensions, and was particularly limited for medical advice, amount, and syndrome. Physicians showed higher within-group consistency than LLMs in categories such as Western diagnosis, usage, and TCM diagnosis, whereas LLMs exhibited relatively higher within-group consistency in syndrome, ingredients, and prescription. Notably, even when cross-group alignment was relatively high, as in TCM diagnosis and ingredients, it remained below the strongest within-group reference. Collectively, these findings indicate that current LLMs can achieve strong overall diagnostic performance while still following prescription strategies that remain only partially aligned with physician practice.

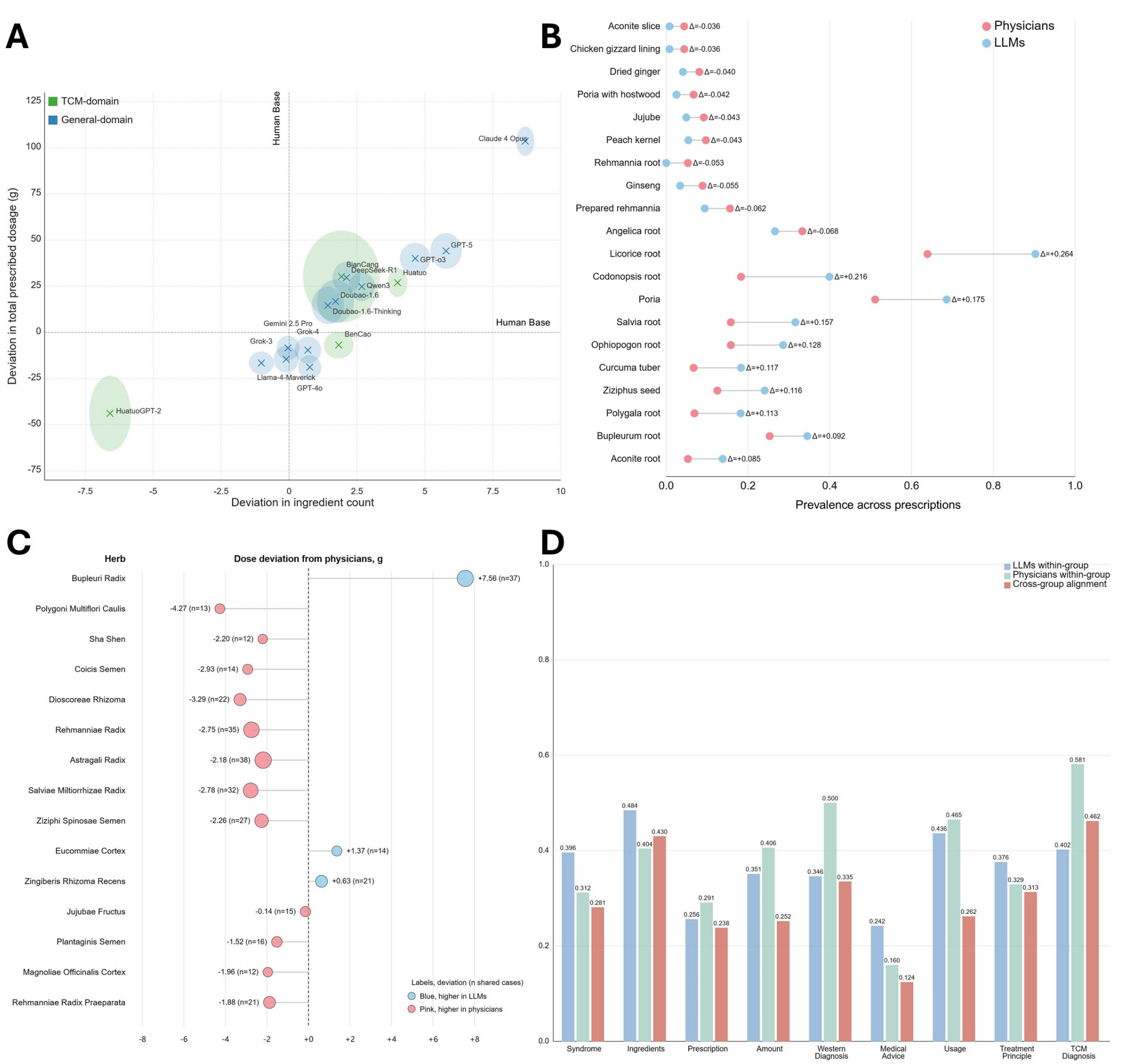

**Fig. 4 | Divergence between LLM-generated and physician prescriptions across composition, dosage, and agreement metrics. (A)** Mean deviation of each model from the

physician case mean in the number of prescribed ingredients (x-axis) and total prescribed dosage (y-axis) across cases. Points indicate model-level mean deviations; shaded ellipses denote 95% confidence intervals. Models closer to the origin more closely resemble physician prescribing patterns. **(B)** Differential prevalence of selected normalized herbs in prescriptions from physicians and LLMs. Each pair of points shows the proportion of prescriptions containing the same herb in the two groups; larger separations indicate stronger preference differences. **(C)** Mean dose deviation from physicians for shared herbs that were prescribed by both groups in at least 10 cases. Values are shown as LLMs minus physicians; blue indicates higher mean dose in LLMs and pink indicates higher mean dose in physicians. Labels denote the mean deviation, with n indicating the number of shared cases. **(D)** Within-group consistency among LLMs, within-group consistency among physicians, and cross-group alignment across nine diagnosis- and prescription-related categories. Higher values indicate greater similarity.

**LLMs offer substantial efficiency gains while resource consumption varies across models**

Resource consumption varied substantially across LLMs, whereas response generation remained consistently faster than that of physicians. **Fig. 5A** shows that input token usage differed appreciably across models, with Claude 4 Opus exhibiting the highest input token counts, whereas DeepSeek-R1 and Doubao-1.6-Thinking generally used more compact inputs. This heterogeneity became even more pronounced in **Fig. 5B**, where GPT-5 produced the longest outputs by a clear margin, and Doubao-1.6 together with Doubao-1.6-Thinking also generated comparatively verbose responses, while Claude Opus 4, Gemini 2.5 Pro, GPT-4o, Grok 4 and Llama-4-Maverick remained relatively concise. As illustrated in **Fig. 5C** through **Fig. 5E**, these differences translated into marked disparities in monetary cost. Claude Opus 4 and GPT-5 incurred the highest total token expenditures overall, GPT-o3 occupied an intermediate tier, and DeepSeek-R1, Doubao-1.6, Llama-4-Maverick and Qwen3-256B remained substantially less expensive. These results indicate that total computational cost was jointly shaped by response length and model-specific pricing rather than by token count alone. In contrast, **Fig. 5F** shows that all evaluated LLMs generated responses substantially faster than physicians. Most models completed inference within seconds to approximately one minute, whereas physician response times extended from minutes to hours and showed considerably greater variability. Together, these findings suggest that LLMs can deliver major gains in turnaround time for clinical decision support, although such efficiency advantages are accompanied by substantial trade-offs in token consumption and economic cost across models.

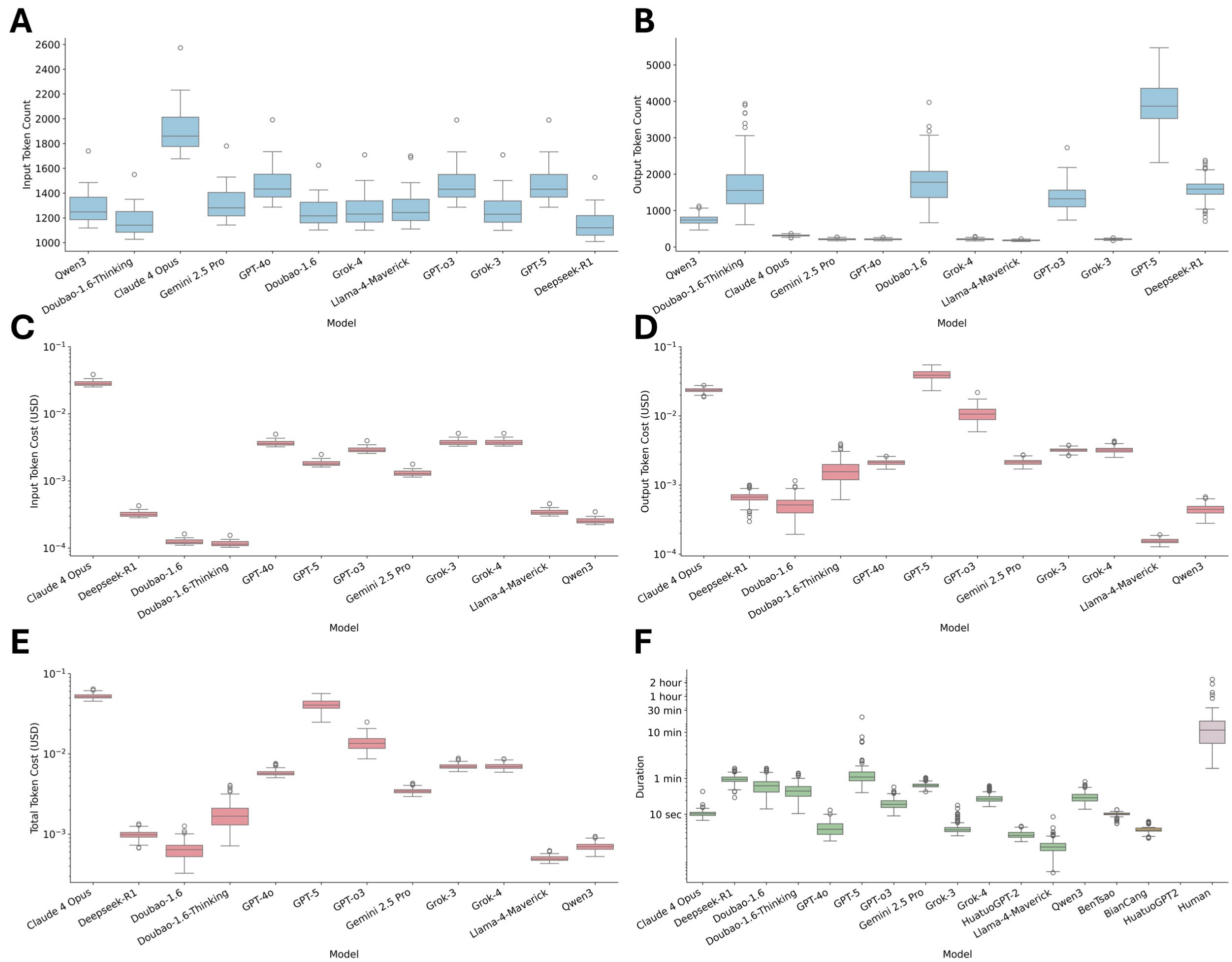

**Fig. 5 | Token usage, monetary cost, and response duration of LLMs compared with physicians. (A)** Distribution of input token counts across evaluated cases for each LLM. **(B)** Distribution of output token counts across evaluated cases for each LLM. **(C)** Input token cost per case (USD) for each LLM. **(D)** Output token cost per case (USD) for each LLM. **(E)** Total token cost per case (USD), combining input and output costs. **(F)** Response duration for each model and for physicians. The y-axis is shown on a logarithmic scale.

### Expert ratings show high reliability across evaluation dimensions

As shown in **Table 2**, inter-rater reliability analysis demonstrated consistently high agreement among the five expert raters across all evaluation dimensions, with overall ICC(2, k) values ranging from 0.748 to 0.940. Agreement was strongest for Western Diagnosis (0.940) and TCM Diagnosis (0.931), followed by Medical Advice (0.904) and Dosage (0.901), indicating that the scoring rubric yielded highly reproducible judgments in core diagnostic and treatment-related domains. By contrast, Usage showed the lowest overall reliability (0.748), suggesting greater subjectivity in evaluating administration-related recommendations. When stratified by source, ratings for AI model outputs were generally more consistent than those for doctors, with higher ICC values in most dimensions, including Western Diagnosis (0.957 vs 0.824), Treatment Principle (0.922 vs 0.712), Ingredients (0.908 vs 0.667), and Prescription (0.903 vs 0.668). The two groups showed near-identical agreement in Dosage (0.901 vs 0.897), whereas both exhibited relatively lower consistency in Usage (0.645 vs 0.643). Taken together, these results indicate that expert scoring was highly reliable overall.

**Table 2. Inter-rater reliability across evaluation dimensions as measured by ICC(2, k)**

| Dimension | Overall ICC(2, k) | AI Models ICC(2, k) | Doctors ICC(2, k) |
|---|---|---|---|
| Western Diagnosis | 0.94 | 0.957 | 0.824 |
| TCM Diagnosis | 0.931 | 0.945 | 0.847 |
| Medical Advice | 0.904 | 0.926 | 0.766 |
| Dosage | 0.901 | 0.901 | 0.897 |
| Treatment Principle | 0.894 | 0.922 | 0.712 |
| Syndrome | 0.892 | 0.913 | 0.785 |
| Ingredients | 0.876 | 0.908 | 0.667 |
| Prescription | 0.87 | 0.903 | 0.668 |
| Usage | 0.748 | 0.645 | 0.643 |

**Hallucination and safety concerns of LLMs**

Hallucinations remained common in both general-purpose and TCM-specialized LLMs, but their forms differed systematically. In general-purpose LLMs, hallucinations mainly appeared as case expansion, whereby the original clinical trajectory was preserved but additional diagnoses, comorbidities, or management details were introduced. For example, in Case 27, physician responses focused primarily on chronic nephritis, hematuria, chronic renal insufficiency, and hypertensive renal injury, whereas GPT-5 additionally introduced chronic kidney disease staging, atrial fibrillation, hyperuricemia, and anticoagulation-related history, and GPT-o3 further added dietary recommendations, renal function follow-up, and anticoagulation management. In Case 14, physician responses centered on cough, throat obstruction, upper respiratory tract infection, or acute pharyngitis, whereas Gemini 2.5 Pro additionally introduced insomnia or sleep disturbance and added fried *Ziziphi Spinosae Semen* to the prescription, thereby extending the therapeutic focus from cough and throat relief to sedation-related management. In Case 21, Doubao-1.6-Thinking supplemented the main presentation of palpitations and arrhythmia with additional findings including left ventricular enlargement, aortic valve calcification with mild regurgitation, and chronic gastritis.

By contrast, hallucinations in TCM-specialized LLMs were more often characterized by cross-case templating and disease substitution. Huatuo repeatedly generated the same diagnostic axis and highly similar prescriptions across distinct cases, including Case 27 (renal disease/oedema), Case 26 (goitre/thyroiditis), Case 14 (cough/throat obstruction) and Case 4 (delayed menstruation/polycystic ovary syndrome). Among the 60 responses generated by BianCang, 42 assigned the TCM diagnosis "*Feiwei disease*" and 27 used "*Feiwei decoction*" as the prescription; accordingly, the renal case in Case 27 and the postoperative metastatic breast cancer case in Case 5 were both misdiagnosed as "*Feiwei disease*", whereas Case 4 was misdiagnosed as "common cold" and treated with Mahuang decoction. Similarly, HuatuoGPT-2 exhibited a highly repetitive output pattern, with "*Tianma Gouteng decoction*" appearing in 32 of 60 responses, "*acute icteric hepatitis*" appearing as the Western diagnosis in 17 cases, and "*migraine*" in 13 cases. As a result, Case 26 (subacute thyroiditis) and Case 14 (cough/acute pharyngitis) were misdiagnosed as migraine, and Case 4 (delayed menstruation/polycystic ovary syndrome) was misdiagnosed as acute icteric hepatitis.

# Discussion

This study systematically evaluated the performance of cutting-edge general-purpose LLMs, TCM-specialized LLMs, and practicing clinicians across multidimensional diagnostic and therapeutic tasks using real-world TCM outpatient cases. By conducting blinded expert evaluation on 60 representative cases, and integrating rank-based comparisons, win–tie–loss analysis, case-level assessment, prescription composition and dosage consistency analysis, as well as efficiency and cost evaluation, we found that state-of-the-art general-purpose LLMs have reached or exceeded the average performance of the physician comparator under blinded expert review across several key dimensions, particularly in medical advice, treatment principles, and selected diagnostic tasks. At the same time, substantial heterogeneity was observed across models: cutting-edge general-purpose models consistently achieved the strongest overall performance, whereas TCM-specialized models did not demonstrate systematic advantages over their general-purpose counterparts, with some exhibiting clear issues such as template-driven outputs, disease substitution, and repetitive prescriptions. Importantly, further analysis revealed that high scores under blinded expert evaluation did not necessarily imply alignment with real-world clinical practice. Model-generated prescriptions showed notable discrepancies from physician practice in terms of herb composition, ingredient selection, dosage control, and overall treatment strategy. For example, LLM-generated prescriptions occasionally included expensive herbal medicines such as ginseng (*Panax ginseng*) and dendrobium (*Dendrobium spp.*), or even substances derived from protected or endangered animals, such as pangolin scales or antelope horn. In addition, LLMs tended to include a larger number of auxiliary ingredients and higher overall dosages, whereas physicians typically balanced therapeutic efficacy with patient affordability, producing more concise and cost-conscious prescriptions.

Differences were also observed in clinical reasoning and management strategies. Physicians often provided initial treatment recommendations while explicitly incorporating follow-up plans, adjusting prescriptions based on patient response, and recommending further examinations in complex cases. In contrast, LLM outputs tended to present more definitive and static conclusions, with limited capacity to integrate multiple coexisting conditions into a unified diagnostic framework. Instead, model-generated diagnoses more frequently resembled a concatenation of independent symptom-based judgments rather than a holistic syndrome-based assessment. Taken together, these findings indicate that improvements in overall performance metrics do not necessarily translate into full alignment with clinical reasoning processes or prescription practices. While LLMs demonstrate strong capabilities in structured reasoning and information synthesis, substantial gaps remain in their ability to replicate the nuanced, context-aware decision-making of clinicians.

Distinct stylistic and formatting characteristics were also observed across general-purpose LLMs, and these differences may further influence their usability and appropriateness in clinical settings. Overall, many models tended to include explanatory or interpretive content beyond the concise, result-oriented style typically used in physician-authored clinical documentation. For example, ChatGPT frequently supplemented diagnoses and prescriptions with parenthetical explanations and occasionally annotated processing methods for individual herbal ingredients; its Western medical diagnoses were comparatively cautious, often including qualifiers such as "requiring further evaluation," and its prescriptions were commonly labeled using classical formula names, with medical advice presented as extended free-text paragraphs. Gemini similarly tended to provide explanatory rationale in parentheses and often appended cautionary statements such as "for reference only" in treatment recommendations; as a reasoning-oriented model, it occasionally produced mixed Chinese–English output or residual English reasoning traces despite Chinese prompts, and in selected cases additionally offered dosage ranges, substitute herbs, or severity alerts. Qwen3-256B also frequently used parenthetical explanations and, in longer interaction contexts, occasionally generated excessively verbose medical advice, suggesting limitations in long-context consistency; it often employed self-defined prescription names with "modified formula" annotations and sometimes produced explicit risk stratification or uncertainty markers in its outputs. DeepSeek was characterized by particularly dense explanatory content, frequent use of technical terminology, and mixed Chinese–English medical expressions; its prescriptions often included extensive processing instructions for herbal components and highly detailed administration guidance, while its diagnostic outputs were broad and frequently included multiple tentative diagnoses pending further work-up. Claude demonstrated a tendency to supplement prescriptions with additional operational details, such as urgency notes in dispensing instructions or preparation annotations for individual ingredients. Collectively, these findings indicate that even among models with comparable quantitative performance, substantial differences remain in output style, explanatory granularity, risk communication, prescription formatting, and patient readability. Such differences suggest that the practical utility of LLMs in clinical deployment depends not only on diagnostic and prescribing performance, but also on whether their outputs conform to clinical documentation norms, facilitate physician review, and are appropriate for downstream patient-facing use.

This study has several limitations. First, although the dataset was constructed from real-world outpatient cases across multiple hospitals, the overall sample size remains limited. In addition, both the selected cases and participating physicians may reflect regional and population-specific characteristics, which could introduce potential biases and limit the generalizability of the findings. Second, although expert scoring based on unified rubric demonstrated high inter-rater reliability, both TCM diagnosis and expert evaluation inherently involve a degree of subjectivity. Such subjectivity may be particularly pronounced in dimensions related to syndrome differentiation, prescription details, and treatment interpretation. Third, the present evaluation was conducted based on static case descriptions rather than dynamic clinical workflows. In real-world practice, physicians iteratively refine their decisions through patient questioning, physical examination, longitudinal follow-up, and experiential judgment. These dynamic processes were not fully captured in the current evaluation framework, and therefore, model performance in this study may not directly translate to actual clinical settings. Fourth, model outputs are influenced by multiple factors, including model version updates, system prompts, API configurations, and cost-related constraints. As a result, the findings reported here represent performance at a specific time point under particular experimental conditions, and may evolve as models continue to be updated. Finally, although we provided illustrative analyses of hallucinations, template-driven outputs, and disease misdiagnosis, these observations were primarily qualitative. A more systematic taxonomy and quantitative framework for characterizing model errors will be necessary in future work to better define the safety boundaries of LLMs in clinical applications.

Despite these limitations, our findings highlight the considerable potential of LLMs to support TCM clinical practice. First, the marked advantage of LLMs in response speed suggests clear utility for high-efficiency preprocessing tasks, including preliminary information organization, extraction of diagnostic highlights, and drafting of structured clinical notes, thereby reducing the burden of repetitive documentation for physicians. Second, the strong reasoning performance of cutting-edge general-purpose LLMs across multiple clinical dimensions indicates that these models may serve as useful reference tools for generating differential diagnoses, summarizing treatment principles, identifying potentially overlooked findings, and offering a second perspective on complex cases. Third, in TCM education and training settings, LLMs may also facilitate standardized case organization, retrieval of classical formula knowledge, clinical teaching, and cross-case pattern induction, thereby supporting the training of junior physicians and improving consistency in clinical reasoning. Fourth, from the perspective of resource utilization, although token consumption and monetary cost varied substantially across models, some systems showed a favorable balance between efficiency and cost, suggesting potential value in primary care, teleconsultation, pre-consultation triage, and other resource-constrained settings. At the same time, our results underscore several major challenges that must be addressed before broader clinical adoption can be considered. Hallucinations remained common in both general-purpose and TCM-specialized models, including additive case expansion and clinically meaningful misdiagnosis, both of which could directly affect diagnostic reasoning and downstream treatment decisions.

In addition, although TCM-oriented LLMs have developed rapidly in recent years, our findings suggest that high-performing TCM-specialized models remain limited, and that current domain-specific systems do not yet consistently surpass strong general-purpose models. Moreover, even when models achieved relatively high scores in expert evaluations, important deficiencies persisted in prescription generation. Divergences in herb composition, dose allocation, and treatment strategy indicate that current models still do not adequately reflect real-world clinical practice, particularly with respect to dosage control, safety margins, and individualized adjustment. These limitations suggest that any near-term clinical use of LLM-generated prescriptions would require strict physician oversight. Practical issues such as model-dependent differences in computational cost and resource consumption may further affect deployment in real healthcare systems. More importantly, the clinical use of LLMs also raises broader questions regarding legal regulation, accountability, and compliance, especially when model outputs may directly influence patient care. These issues are particularly salient in high-risk tasks such as prescription generation.

Collectively, these findings suggest that substantial work remains to improve the reliability, faithfulness, and controllability of LLM-generated clinical outputs for TCM. This includes reducing hallucinations, strengthening alignment with source clinical information, and incorporating explicit safety constraints into prescription generation. The development of more clinically adapted TCM-specialized models remains an important direction, but their progress should be evaluated against strong general-purpose baselines rather than assumed a priori. Equally important will be prospective studies embedded in real clinical workflows to determine whether LLM-assisted systems can safely and effectively improve diagnostic efficiency, documentation quality, and patient management under physician supervision. More systematic error taxonomies, safety benchmarks, and post-deployment monitoring frameworks will also be needed to define the boundaries of acceptable use in clinical settings. In summary, this study shows that LLMs have already demonstrated meaningful potential for integration into clinical decision-support systems, but their translation into safe, reliable, and scalable clinical practice will require more rigorous evaluation, stronger model alignment, and deeper integration with physician-centered workflows.

# Methods

### Real-world clinical case collection and benchmark construction

We collected 349 real-world clinical cases from 75 physicians across China, spanning 2018 to 2023. From this cohort, we further curated a subset of 60 representative cases to construct the benchmark dataset for evaluation. This subset comprised 41 females (68.3%) and 19 males (31.7%) patients, with a broad age distribution (mean age 55.5 years, median 56 years, range 26–83 years). Older adults were well represented, with 60–69 years accounting for 21.7% (n = 13) and ≥70 years for 23.3% (n = 14). In terms of disease composition, the selected cases covered a diverse spectrum of clinical conditions, including renal/urinary disorders (n = 10, 16.7%), thyroid-related diseases (n = 9, 15.0%), gynecological/reproductive conditions (n = 6, 10.0%), metabolic disorders (n = 6, 10.0%), cardiovascular diseases (n = 6, 10.0%), respiratory diseases (n = 5, 8.3%), liver and cirrhosis-related conditions (n = 5, 8.3%), digestive diseases (n = 4, 6.7%), and dizziness/vertigo-related conditions (n = 3, 5.0%), thereby capturing both common outpatient presentations and clinically complex cases. The benchmark cases were contributed by 12 physicians, with initial visits concentrated between 2021 and 2023. Notably, these cases are recent and have not been previously published, reducing the likelihood that they were included in the pretraining corpora of contemporary LLMs and enabling a more objective assessment of model generalization in real-world TCM clinical scenarios.

At the level of case design, each case preserves a comprehensive and structured representation of the clinical encounter, closely reflecting real-world outpatient records. Specifically, each case includes demographic information (e.g., sex, age, occupation), initial consultation date, chief complaint, history of present illness, and detailed clinical presentation, together with key elements of the four diagnostic methods in TCM (inspection, auscultation/olfaction, inquiry, and palpation), including mental status, appetite, sleep, bowel and urinary function, tongue characteristics, and pulse findings. In addition, cases incorporate physical examination findings and a wide range of auxiliary diagnostic data, such as imaging studies, laboratory tests, pathological reports, and immunohistochemical results, providing objective clinical evidence alongside narrative descriptions.

Beyond basic clinical information, each case systematically documents both TCM and Western medical diagnoses, syndrome differentiation, treatment principles, herbal prescriptions (including individual herbs and dosages), preparation and administration methods, prescription duration, and detailed medical advice, thereby enabling a comprehensive evaluation of diagnostic reasoning, syndrome differentiation, and prescription generation.

Furthermore, a subset of cases includes longitudinal follow-up information, such as second and subsequent consultations, documenting symptom progression, treatment response, and prescription adjustments. This allows the dataset to capture not only initial decision-making but also elements of dynamic clinical management. For selected representative cases, we additionally retained the treating physician's clinical commentary, which provides explicit explanations of pathophysiological reasoning, treatment strategies, and prescription modifications. For example, in a case of gastric cancer during chemotherapy treated with a spleen-strengthening and qi-tonifying formula, the record includes comprehensive imaging, pathology, and immunohistochemical findings, integrated TCM and Western diagnoses, detailed herbal prescriptions with dosage adjustments across follow-up visits, and a structured exposition of the underlying syndrome differentiation and therapeutic rationale.

**Recruitment of participating clinicians**

We recruited 67 frontline TCM clinicians from 47 tertiary hospitals across China to participate in the physician evaluation arm of this study, and 60 physicians were ultimately retained as the final human comparator cohort after quality control screening for response completeness and formatting compliance. **Fig. 6A** illustrates the case-selection preferences of participating physicians across the benchmark cases. As shown in **Fig. 6B**, the participating clinicians spanned a broad range of clinical experience, with a mean practice duration of 13.47 years, from early-career to highly experienced physicians.

As summarized in **Fig. 6C,** intermediate-level physicians constituted the largest subgroup (n = 32, 53.3%)**,** while 17 clinicians (28.3%) held senior professional titles and 8 clinicians (13.3%) were junior physicians. As shown in **Fig. 6D**, the final physician cohort included 38 female clinicians (63.3%) and 22 male clinicians (36.7%). The participating clinicians also represented diverse clinical specialties. As illustrated in **Fig. 6E**, the largest proportion came from general TCM practice (n = 22), followed by pediatrics (n = 11), Tuina (n = 10**)**, acupuncture (n = 8)**,** gastroenterology (n = 3), and otorhinolaryngology (n = 2), with additional representation from pulmonology, neurology, endocrinology, and brain disorders.

To better approximate real-world outpatient practice, participating clinicians were allowed to preferentially select cases aligned with their subspecialty expertise whenever possible. Each physician independently reviewed and diagnosed the assigned benchmark cases under standardized formatting instructions identical to those provided to the evaluated LLMs.

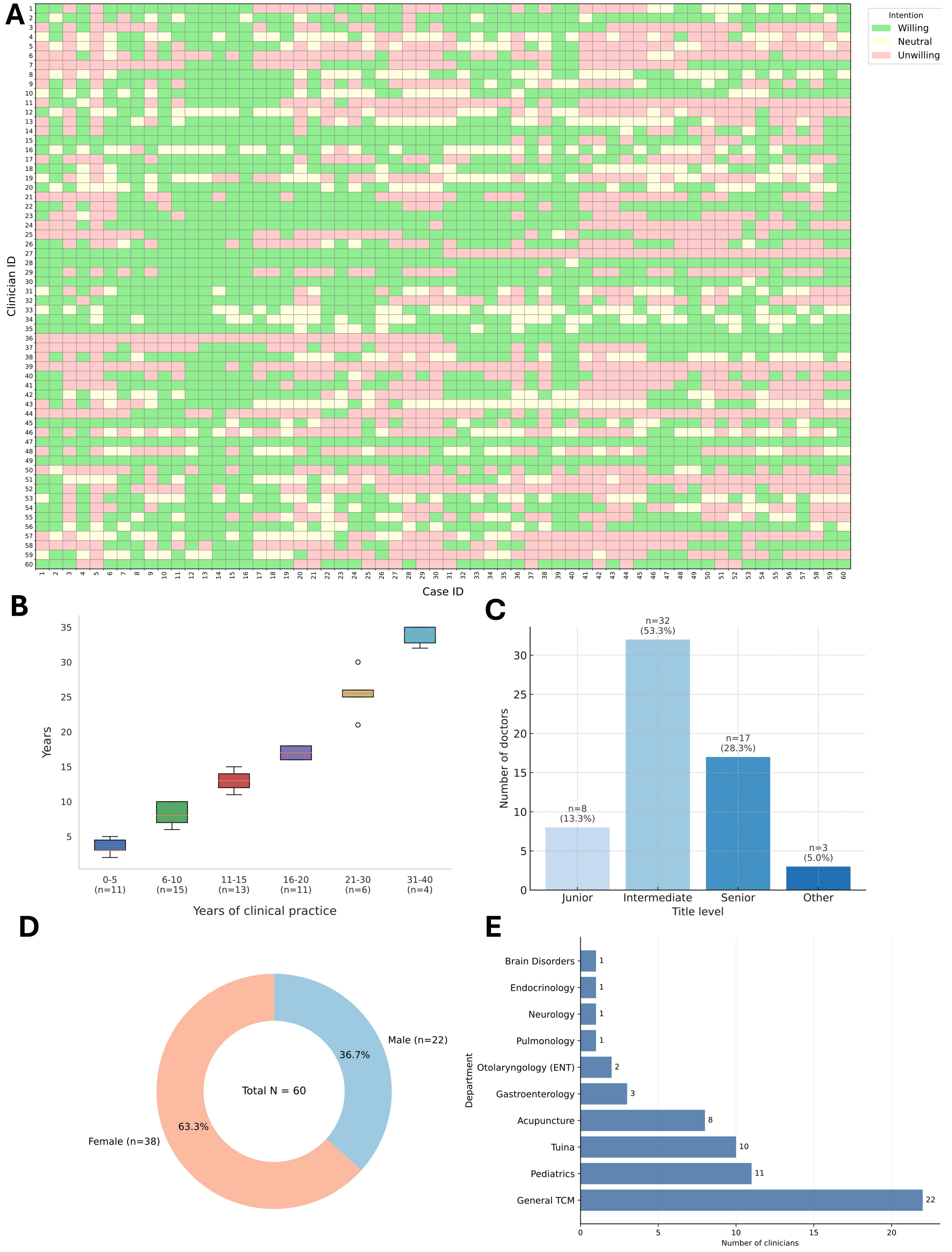
A
Intention
Willing
Neutral
Unwilling
Clinician ID
Case ID
B
Years
Years of clinical practice
0-5 (n=11)
6-10 (n=15)
11-15 (n=13)
16-20 (n=11)
21-30 (n=6)
31-40 (n=4)
C
Number of doctors
n=8 (13.3%)
n=32 (53.3%)
n=17 (28.3%)
n=3 (5.0%)
Junior
Intermediate
Senior
Other
Title level
D
Male (n=22)
36.7%
Total N = 60
63.3%
Female (n=38)
E
Department
Brain Disorders
Endocrinology
Neurology
Pulmonology
Otolaryngology (ENT)
Gastroenterology
Acupuncture
Tuina
Pediatrics
General TCM
Number of clinicians

**Fig. 6 | Characteristics of participating clinicians and case selection patterns in the physician evaluation cohort. (A)** Heatmap showing physicians' willingness to assign cases across the 60 benchmark clinical cases. Each row represents an individual clinician and each column represents a benchmark case. Green indicates cases in which the clinician was willing to evaluate, pink indicates cases in which the clinician was unwilling, and beige indicates neutral preference. **(B)** Distribution of clinical practice duration among participating physicians, stratified by years of clinical experience. Box plots show median, interquartile range, and outliers. **(C)** Distribution of professional title levels among participating clinicians. **(D)** Sex distribution of the final physician cohort. **(E)** Clinical specialty distribution of participating physicians, showing representation across general TCM and subspecialty departments.

### LLM selection and setting

To comprehensively benchmark LLM performance in TCM clinical reasoning, we evaluated 12 general-purpose LLMs and 4 TCM-specialized LLMs, selected to represent the strongest publicly accessible systems available during the study period. The general-purpose models included DeepSeek-R1, GPT-o3, Claude Opus 4, Qwen3-256B, Grok 3, Doubao-1.6-Thinking, Doubao-1.6, Llama-4-Maverick, GPT-5, Grok 4, GPT-4o, and Gemini 2.5 Pro. These models were chosen to cover a broad spectrum of leading cutting-edge systems from major commercial and open-model ecosystems, with emphasis on strong reasoning and instruction-following capabilities. The TCM-specialized models included BenCao[46], Huatuo[47], HuatuoGPT-2[45], and BianCang[44], all of which were specifically developed or adapted for Traditional Chinese Medicine–related tasks. These domain-specific models were selected based on their public availability, reproducible inference, and representativeness within the current TCM LLM landscape. In addition to the benchmarked TCM-specific models, we surveyed several other TCM-oriented LLMs reported in the literature or public repositories. However, some were excluded from final evaluation because their model weights were not publicly released, inference services were unavailable, or deployment conditions could not be reliably reproduced at the time of study.

### Experimental procedure and expert evaluation

All benchmark LLMs evaluated the same set of 60 representative real-world TCM outpatient cases under a standardized assessment framework. Physicians evaluated assigned case subsets based on subspecialty preferences, such that each physician evaluated 6 cases and each case was independently evaluated by 6 physicians. For each case, the complete de-identified clinical record was provided as input, including demographic information, chief complaint, present illness history, relevant findings from the four diagnostic methods of TCM, tongue and pulse descriptions, specialist examinations, and auxiliary laboratory or imaging results. Each participating LLM received an identical prompt requesting structured outputs across multiple clinical dimensions, including TCM diagnosis, syndrome differentiation, Western diagnosis, treatment principle, prescription name, herbal ingredients, dosage, usage instructions, and medical advice. Physicians were instructed to respond using the same output schema to maximize comparability across groups. The full prompting template used in this study is provided in **Supplementary Table 3**.

Model inference was conducted independently on a case-by-case basis. For proprietary models, responses were obtained through official APIs or publicly accessible interfaces; for open-weight models, inference was performed using reproducible local or hosted deployment pipelines. Unless otherwise specified, default provider-recommended generation settings were adopted. For each evaluated LLM, we recorded the model version, access date, provider or deployment mode, inference interface, generation settings, system prompt, maximum output length, web-access status, number of repeated runs, and pricing information. Detailed model configurations, prompting templates, and evaluation settings are publicly available through the project repository described in **Code availability**. Physicians independently completed their assigned cases without access to model outputs or other physicians' responses. To better approximate routine clinical practice, physicians were allowed to preferentially select cases aligned with their subspecialty expertise whenever feasible.

All responses were subsequently assessed through a single-blind expert review process. Five senior TCM clinicians with substantial clinical experience served as independent raters and were blinded to the source of each response (LLM or physician). For each case, outputs were anonymized and randomly reordered before scoring. Expert scoring was conducted through a dedicated web-based evaluation platform developed specifically for this study, accessible at [http://app.itongue.cn:8123/](http://app.itongue.cn:8123/). The platform enabled structured side-by-side review, standardized score entry, and centralized data collection. Representative screenshots of the scoring interface are provided in **Supplementary Figure 1**.

Evaluators scored responses across nine predefined dimensions: TCM Diagnosis, Syndrome, Western Diagnosis, Treatment Principle, Prescription, Ingredients, Amount, Usage, and Medical Advice. Each dimension was rated using a five-point Likert scale, where higher scores indicated closer alignment with expert expectations and real-world clinical practice. Scoring was performed according to a unified rubric developed before the study, with reference to both the original patient records and the diagnoses and prescriptions used in actual clinical care. The complete scoring criteria are provided in **Supplementary Table 2**.

To improve robustness, final scores were aggregated across expert raters, and inter-rater reliability was quantified using intraclass correlation coefficients. Comparative analyses included dimension-level score comparisons, overall ranking analyses, win–tie–loss comparisons against physicians, case-level performance comparisons, prescription composition and dosage consistency analyses, and resource-efficiency assessments including token consumption, monetary cost, and response time.

In addition to the primary model–physician comparison, we conducted an exploratory clinician–AI interaction analysis to assess whether LLM-generated support could improve junior physicians' diagnostic performance. Seven junior TCM physicians independently evaluated the benchmark cases before and after access to DeepSeek-generated diagnostic support. In the unaided phase, physicians generated structured diagnostic reports using the same output schema as the main evaluation. In the assisted phase, physicians reviewed the same case materials together with DeepSeek-generated diagnostic suggestions and then produced revised structured reports. Both unaided and assisted reports were anonymized, pooled with other study outputs, and evaluated by the same five senior TCM experts using the predefined nine-dimensional scoring rubric. The overall score for each report was calculated by averaging scores across the nine evaluation dimensions and expert raters; score improvement was defined as the paired difference between assisted and unaided reports for each junior physician.

**Statistical analysis**

Overall differences among models were assessed using the Friedman test, treating the benchmark case as the blocking factor and the model as the comparison factor. When the global test indicated evidence of performance differences, post hoc pairwise comparisons were performed using the Nemenyi procedure[48]. Mean ranks were used to summarize relative model performance across cases, and statistically comparable groups were visualized using critical difference diagrams, following established recommendations for comparing multiple algorithms across matched datasets[48]. Within each clinical dimension, the same rank-based framework was applied to assess whether relative model performance differed across cases. For non-matched distributional comparisons, such as selected prescription-level summary metrics where the paired-case structure was not applicable, the Kruskal–Wallis test was used[49].

For direct comparisons with the physician cohort, model performance was evaluated under a paired-case framework. For each benchmark case and clinical dimension, the physician reference was defined as the mean score of practicing TCM physicians for the same case and dimension. Model–physician score differences were then calculated at the case level. Paired differences were tested using two-sided Wilcoxon signed-rank tests, with Benjamini–Hochberg correction applied across multiple model–dimension contrasts[50,51]. In parallel, win–tie–loss analyses were used to provide an interpretable case-level summary: each case was classified as a model win, tie or loss according to whether the model score was higher than, equal to or lower than the corresponding physician reference. Additional case-level analyses quantified the number of individual physicians each model outperformed across benchmark cases.

As a sensitivity analysis, repeated-case score data were modeled using linear mixed-effects models. Model identity, clinical dimension and their interaction were included as fixed effects, and case ID was included as a random intercept to account for within-case correlation across model outputs[52]. Estimated marginal means were used to compare each model with the physician reference, with Benjamini–Hochberg correction applied to multiple contrasts[51]. These models were used to assess whether the main rank-based findings were robust to a parametric repeated-measures specification.

Agreement among expert raters was assessed using the two-way random-effects intraclass correlation coefficient for averaged ratings, ICC(2,k), because each output was evaluated by the same panel of raters and the raters were considered representative of a broader expert population[53,54]. ICC estimates were reported with 95% confidence intervals where applicable.

Prescription-related analyses compared model-generated prescriptions with physician references. We examined deviations in ingredient count and total dosage, herb prevalence distributions, dose deviations among shared herbs, and within-group versus between-group consistency in prescription composition. For shared-herb analyses, dose deviation was calculated only for herbs appearing in both the model-generated and physician-reference prescriptions. Text similarity analyses for selected diagnostic outputs were performed using sentence-transformers embeddings and cosine similarity, following the sentence-BERT framework for deriving semantically meaningful sentence embeddings[55].

Resource-efficiency analyses summarized input-token count, output-token count, total token expenditure, estimated monetary cost and response generation time for each model. Cost estimates were calculated using model-specific unit prices recorded at the time of evaluation. Unless otherwise specified, continuous variables were summarized using means with standard deviations or medians with interquartile ranges, as appropriate. All statistical tests were two-sided. P values were adjusted for multiple comparisons where applicable, and adjusted $P < 0.05$ was considered statistically significant unless otherwise stated. Analyses were performed in Python using pandas (v2.2.3), NumPy (v1.26.4), SciPy (v1.15.2), scikit-learn (v1.6.1), sentence-transformers, PyTorch (v2.1.2), and transformers (v4.49.0). Visualization was conducted using Matplotlib (v3.10.0) and Seaborn (v0.13.2).

**Ethical approval**

The study protocol was reviewed and approved by the IRB of Wanggang Community Health Service Center, Pudong New Area, Shanghai (approval number: 2025-001). The approved study documents included the clinical research protocol, informed consent form, recruitment materials, case report form, subject identification code table, and related investigator qualification materials. All clinical data were de-identified before analysis. Written informed consent was obtained from all patients whose clinical data were included and from all participating physicians, in accordance with the approved protocol.

## Data availability

De-identified data supporting the findings of this study will be made available upon publication where permitted by ethics approval and participant consent. The 60 de-identified clinical cases used for evaluation are included in the **Supplementary Materials**. Additional de-identified clinical data may be available from the corresponding authors upon reasonable request and completion of an ethics-compliant data-use agreement.

## Code availability

Code to reproduce the main and supplementary analyses is available via GitHub at https://github.com/orangeshushu/TCM-Evaluation. The repository also contains the complete evaluation framework, prompting templates, model versions, access dates, deployment modes, inference interfaces, generation settings, web-access status, repeated-run configurations, and statistical analysis scripts used in this study, enabling transparent and reproducible benchmarking of both physicians and large language models.

## Acknowledgements

This work is partially supported by the Paul K. and Diane Shumaker Endowment Fund to Dong Xu at the University of Missouri. The authors would like to express sincere gratitude to the following physicians, who are acknowledged here by name with their permission, for voluntarily participating in this study and for providing essential diagnostic information for the relevant cases: Shichun Guan (Jinzhai County Hospital of Traditional Chinese Medicine), Lijuan Liu (Jinsha County Hospital of Traditional Chinese Medicine), Jinghui Shi (Jiaxing Road Subdistrict Community Health Service Center, Hongkou District, Shanghai), Xu Zhou (Yangpu District Hospital of Traditional Chinese Medicine), Yifeng Yu (Shuguang Hospital Affiliated to Shanghai University of Traditional Chinese Medicine), Lian Yao (Jinsha County Hospital of Traditional Chinese Medicine), Xinjun Sun (Qingpu District Hospital of Traditional Chinese Medicine), Ruijuan Zhang (Hebei University of Chinese Medicine), Jing Zhang (Zhang Jing Traditional Chinese Medicine Clinic, Luyang District, Hefei), Jing Zhang (Shanghai University of Traditional Chinese Medicine), Jing Cao (Yubei District Maternal and Child Health Hospital), Yun Cao (Bo'ai Outpatient Department, Nankai District, Tianjin), Ke Zeng (Shuguang Hospital Affiliated to Shanghai University of Traditional Chinese Medicine), Yuzhang Zhu (Guangming

Hospital of Traditional Chinese Medicine, Pudong New Area, Shanghai), Chunxiang Li (Bozhou Hospital of Traditional Chinese Medicine), Chunying Yang (Affiliated Hospital of Jiangsu University), Yan Jiang (Shanghai University of Sport), Tie Shen (Hangtou Hesha Community Health Service Center, Pudong New Area, Shanghai), Limiao Wang (Sanlin Community Health Service Center), Dexiang Wang (Qiandongnan Prefecture People's Hospital, Guizhou Province), Qian Wang (Jiading District Hospital of Traditional Chinese Medicine, Shanghai), Qixiang Cheng (Chedun Town Community Health Service Center, Songjiang District, Shanghai), Bo Cheng (Suwenxuan Traditional Chinese Medicine Clinic), Xiaopeng Hu (Huzhu County People's Hospital), Jie Yuan (Fudan University Hospital), Wenrong Miao (Naval Medical Center), Xia Dong (The First People's Hospital of Lanzhou), Junhao Cai (Shanghai Hospital of Traditional Chinese Medicine), Fulin Yuan (Jinsha County Hospital of Traditional Chinese Medicine), Feng Yuan (Wuhan Hospital of Traditional Chinese Medicine), Tan (Yueyang Hospital), Jixiu Zhao (Huize County Hospital of Traditional Chinese Medicine), Yidi Zhao (The First People's Hospital of Lanzhou), Qing Lang (Beicai Community Health Service Center), Qian Zheng (Tianshan Road Subdistrict Community Health Service Center), Mingfu Zheng (The First People's Hospital of Lanzhou), Danni Chen (Shuguang Hospital Affiliated to Shanghai University of Traditional Chinese Medicine), Hongmei Chen (Jinsha County Hospital of Traditional Chinese Medicine), Di Chen (Anhui Provincial Hospital of Traditional Chinese Medicine), Haotian Han (Shanghai University of Traditional Chinese Medicine), Limin Gao (Pudong New Area Mental Health Center), Guorun Gao (Yangliuqing Town Community Health Service Center, Xiqing District, Tianjin), Hongyan Gao (Shanghai University of Traditional Chinese Medicine), Qing Gao (Shandong Provincial Hospital of Traditional Chinese Medicine), Dandan Qi (Guidu Community Health Station, Suzhou Industrial Park), and Dejing Wei (Jinsha County Hospital of Traditional Chinese Medicine).

## Author contributions

J.X., D.X. and G.A. conceived the study. J.X., Y. Yu., J.L. and Z.Z. designed the methodology. Z.Z. developed the evaluation and scoring system. X.T. and G.A. collected the data. X.T., G.A., S.L., C.J., Y. Yang, Z. Zhao and Q.S. performed diagnostic evaluation. J.X. analyzed the data and drafted the manuscript. J.X., D.X. and Y. Yu. revised the manuscript. All authors reviewed and approved the final manuscript.

# Competing interests

J.X. participated in the development of BenCao, one of the TCM-specific models evaluated in this study. The authors declare no other financial or non-financial competing interests.

## References

1. Glaubitz, R. *et al.* The cost of the diagnostic odyssey of patients with suspected rare diseases. *Orphanet J Rare Dis* **20**, 222 (2025).
2. Palmer, E. E. *et al.* Equity in action: The diagnostic working group of the undiagnosed diseases network international. *Npj Genomic Med.* **9**, 37 (2024).
3. Thirunavukarasu, A. J. *et al.* Large language models in medicine. *Nat. Med.* **29**, 1930–1940 (2023).
4. Lee, P., Bubeck, S. & Petro, J. Benefits, limits, and risks of GPT-4 as an AI chatbot for medicine. *N. Engl. J. Med.* **388**, 1233–1239 (2023).
5. Clusmann, J. *et al.* The future landscape of large language models in medicine. *Commun. Med.* **3**, 141 (2023).
6. Kung, T. H. *et al.* Performance of ChatGPT on USMLE: Potential for AI-assisted medical education using large language models. *Plos Digital Health* **2**, e0000198 (2023).
7. Singhal, K. *et al.* Large language models encode clinical knowledge. *Nature* **620**, 172–180 (2023).
8. Singhal, K. *et al.* Toward expert-level medical question answering with large language models. *Nat. Med.* **31**, 943–950 (2025).
9. Jiang, L. Y. *et al.* Health system-scale language models are all-purpose prediction engines. *Nature* **619**, 357–362 (2023).
10. Chen, S. F. *et al.* LLM-assisted systematic review of large language models in clinical medicine. *Nat. Med.* **32**, 1152–1159 (2026).
11. Agrawal, M., Chen, I. Y., Gulamali, F. & Joshi, S. The evaluation illusion of large language models in medicine. *Npj Digital Med.* **8**, 600 (2025).

12. Hager, P. *et al.* Evaluation and mitigation of the limitations of large language models in clinical decision-making. *Nat. Med.* **30**, 2613–2622 (2024).
13. Cabral, S. *et al.* Clinical reasoning of a generative artificial intelligence model compared with physicians. *Jama Intern. Med.* **184**, 581–583 (2024).
14. Eriksen, A. V., Möller, S. & Ryg, J. Use of GPT-4 to diagnose complex clinical cases. *Nejm Ai* **1**, AIp2300031 (2024).
15. Guo, Y. *et al.* Development and prospective shadow evaluation of a domain-specific large language model for emergency neurological diagnosis. *Npj Digital Med.* https://doi.org/10.1038/s41746-026-02644-z (2026) doi:10.1038/s41746-026-02644-z.
16. Dai, H.-J. *et al.* Evaluating real-world deployment of an HL7-CDA-aligned LLM for ICD-10-CM coding. *Npj Digital Med.* https://doi.org/10.1038/s41746-026-02541-5 (2026) doi:10.1038/s41746-026-02541-5.
17. Ayers, J. W. *et al.* Comparing physician and artificial intelligence chatbot responses to patient questions posted to a public social media forum. *Jama Intern. Med.* **183**, 589–596 (2023).
18. Asgari, E. *et al.* A framework to assess clinical safety and hallucination rates of LLMs for medical text summarisation. *Npj Digital Med.* **8**, 274 (2025).
19. Draelos, R. L. *et al.* Large language models provide unsafe answers to patient-posed medical questions. *Npj Digital Med.* **9**, 241 (2026).
20. Tam, T. Y. C. *et al.* A framework for human evaluation of large language models in healthcare derived from literature review. *Npj Digital Med.* **7**, 258 (2024).
21. Vasey, B. *et al.* Reporting guideline for the early stage clinical evaluation of decision support systems driven by artificial intelligence: DECIDE-AI. *BMJ* **377**, e070904 (2022).

22. Collins, G. S. *et al.* TRIPOD+AI statement: Updated guidance for reporting clinical prediction models that use regression or machine learning methods. *BMJ* **385**, e078378 (2024).
23. WHO traditional medicine strategy: 2014-2023. https://www.who.int/publications/i/item/9789241506096.
24. Global traditional medicine strategy 2025-2034. https://www.who.int/publications/i/item/9789240113176.
25. Jiang, M. *et al.* Syndrome differentiation in modern research of traditional Chinese medicine. *Journal of Ethnopharmacology* **140**, 634–642 (2012).
26. Chung, V. C. H., Ho, R. S. T., Wu, X. & Wu, J. C. Y. Incorporating traditional Chinese medicine syndrome differentiation in randomized trials: Methodological issues. *European Journal of Integrative Medicine* **8**, 898–904 (2016).
27. Zhang, B., Jiang, Y., Cheng, C., Lin, H. & Guo, Y. External application of two unrestricted herbal medicines to treat costochondritis in a young collegiate athlete: A case report. *Journal of Integrative Medicine* **18**, 450–454 (2020).
28. Kang, H. *et al.* Integrating clinical indexes into four-diagnostic information contributes to the traditional Chinese medicine (TCM) syndrome diagnosis of chronic hepatitis B. *Sci. Rep.* **5**, 9395 (2015).
29. Liu, Y. *et al.* Evaluating the role of large language models in traditional Chinese medicine diagnosis and treatment recommendations. *Npj Digital Med.* **8**, 466 (2025).
30. Xie, J. *et al.* TCM-ladder: A benchmark for multimodal question answering on traditional Chinese medicine. Preprint at https://doi.org/10.48550/arXiv.2505.24063 (2025).

31. Wang, B. *et al.* TCMEval-PA: A question-answering benchmark dataset for the prescription audit of traditional Chinese medicine. *Sci Data* **13**, 79 (2025).
32. Wang, Z. *et al.* TCMEval-SDT: A benchmark dataset for syndrome differentiation thought of traditional Chinese medicine. *Sci. Data* **12**, 437 (2025).
33. Singh, A. *et al.* OpenAI GPT-5 system card. Preprint at https://doi.org/10.48550/arXiv.2601.03267 (2026).
34. OpenAI *et al.* GPT-4o system card. Preprint at https://doi.org/10.48550/arXiv.2410.21276 (2024).
35. OpenAI. *OpenAI O3 and O4-Mini System Card*. https://cdn.openai.com/pdf/2221c875-02dc-4789-800b-e7758f3722c1/o3-and-o4-mini-system-card.pdf (2025).
36. Gemini Team. *Gemini 2.5: Pushing the Frontier with Advanced Reasoning, Multimodality, Long Context, and next Generation Agentic Capabilities*. https://storage.googleapis.com/deepmind-media/gemini/gemini_v2_5_report.pdf (2025).
37. xAI. Grok 3 beta: The age of reasoning agents. *xAI* https://x.ai/news/grok-3 (2025).
38. xAI. Grok 4. *xAI* https://x.ai/news/grok-4 (2025).
39. Anthropic. *System Card: Claude Opus 4 & Claude Sonnet 4*. https://assets.anthropic.com/m/6c940a1b69ed6a1c/original/Claude-4-System-Card.pdf (2025).
40. Meta AI. The llama 4 herd: The beginning of a new era of natively multimodal AI innovation. *Meta AI* https://ai.meta.com/blog/llama-4-multimodal-intelligence/ (2025).
41. ByteDance Seed Team. Seed1.6. *ByteDance* https://seed.bytedance.com/en/seed1_6 (2025).
42. Yang, A. *et al.* Qwen3 technical report. Preprint at https://doi.org/10.48550/arXiv.2505.09388 (2025).

43. Guo, D. *et al.* DeepSeek-R1 incentivizes reasoning in LLMs through reinforcement learning. *Nature* **645**, 633–638 (2025).

44. Wei, S. *et al.* BianCang: A traditional Chinese medicine large language model. *IEEE Journal of Biomedical and Health Informatics* 1–12 (2025) doi:10.1109/JBHI.2025.3612415.

45. Chen, J. *et al.* HuatuoGPT-II, one-stage training for medical adaption of LLMs. Preprint at https://doi.org/10.48550/arXiv.2311.09774 (2024).

46. Xie, J. *et al.* BenCao: An instruction-tuned large language model for traditional Chinese medicine. Preprint at https://doi.org/10.48550/arXiv.2510.17415 (2025).

47. Wang, H. *et al.* HuaTuo: Tuning LLaMA model with Chinese medical knowledge. Preprint at https://doi.org/10.48550/arXiv.2304.06975 (2023).

48. Demšar, J. Statistical comparisons of classifiers over multiple data sets. *J. Mach. Learn. Res.* **7**, 1–30 (2006).

49. Kruskal, W. H. & Wallis, W. A. Use of ranks in one-criterion variance analysis. *J. Am. Stat. Assoc.* **47**, 583–621 (1952).

50. Wilcoxon, F. Individual comparisons by ranking methods. *Biometrics Bulletin* **1**, 80–83 (1945).

51. Benjamini, Y. & Hochberg, Y. Controlling the false discovery rate: A practical and powerful approach to multiple testing. *Royal Statistical Society. Journal. Series B: Methodological* **57**, 289–300 (1995).

52. Laird, N. M. & Ware, J. H. Random-effects models for longitudinal data. *Biometrics* **38**, 963–974 (1982).

53. Shrout, P. E. & Fleiss, J. L. Intraclass correlations: Uses in assessing rater reliability. *Psychol. Bull.* **86**, 420–428 (1979).

54. Koo, T. K. & Li, M. Y. A guideline of selecting and reporting intraclass correlation coefficients for reliability research. *Journal of Chiropractic Medicine* **15**, 155–163 (2016).

55. Reimers, N. & Gurevych, I. Sentence-BERT: Sentence embeddings using siamese BERT-networks. in *Proceedings of the 2019 Conference on Empirical Methods in Natural Language Processing and the 9th International Joint Conference on Natural Language Processing (EMNLP-IJCNLP)* (eds Inui, K., Jiang, J., Ng, V. & Wan, X.) 3982–3992 (Association for Computational Linguistics, Hong Kong, China, 2019). doi:10.18653/v1/D19-1410.